\documentclass[review]{elsarticle}
\usepackage{times}

\usepackage{soul}
\usepackage{url}
\usepackage[hidelinks]{hyperref}
\usepackage[utf8]{inputenc}
\usepackage[small]{caption}
\usepackage{graphicx}
\usepackage{amsmath}
\usepackage{amsfonts}
\usepackage{multicol}
\usepackage{multirow}
\usepackage{booktabs}
\usepackage{algorithm}
\usepackage{algorithmic}
\usepackage{hyperref}

\journal{Journal of \LaTeX\ Templates}

\DeclareMathOperator*{\argmax}{arg\,max}

\begin{document}

\begin{frontmatter}

\title{Confidence Conditioned Knowledge Distillation}

\author{Sourav Mishra}
\address{Department of Aerospace Engineering, Indian Institute of Science, Bangalore}

\author{Suresh Sundaram}
\address{Department of Aerospace Engineering, Indian Institute of Science, Bangalore}

%
%

\begin{abstract}
 In this paper, a novel confidence conditioned knowledge distillation (CCKD) scheme for transferring the knowledge from a teacher model to a student model is proposed. Existing state-of-the-art methods employ fixed loss functions for this purpose and ignore the different levels of information that need to be transferred for different samples. In addition to that, these methods are also inefficient in terms of data usage. CCKD addresses these issues by leveraging the confidence assigned by the teacher model to the correct class to devise sample-specific loss functions (CCKD-L formulation) and targets (CCKD-T formulation). Further, CCKD improves the data efficiency by employing self-regulation to stop those samples from participating in the distillation process on which the student model learns faster. Empirical evaluations on several benchmark datasets show that CCKD methods achieve at least as much generalization performance levels as other state-of-the-art methods while being data efficient in the process. Student models trained through CCKD methods do not retain most of the misclassifications commited by the teacher model on the training set. Distillation through CCKD methods improves the resilience of the student models against adversarial attacks compared to the conventional KD method. Experiments show at least $3\%$ increase in performance against adversarial attacks for the MNIST and the Fashion MNIST datasets, and at least $6\%$ increase for the CIFAR10 dataset.
\end{abstract}


\end{frontmatter}


\section{Introduction}
Deep learning models have been successfully applied to several computer vision problems such as classification \cite{im_class}, segmentation, object detection \cite{obj}, etc. However, deploying them on edge devices such as mobile phones, drones, etc., is not feasible due to their larger memory footprint \cite{normal}. Therefore several methods have been proposed in the literature to address model compression while retaining most of the generalization power of the original model. These methods can be broadly categorized into two groups based on their assumption about knowledge representation, namely, model compression-based methods and knowledge distillation-based methods \cite{review}.

Model compression-based methods assume that the knowledge is present in the weights of the model \cite{Han2016}. LeCun introduced pruning the neural network without compromising the performance in ~\cite{brainDamage}. Various other methods for compressing neural networks have been proposed in the literature ~\cite{Han2015,Wang2016,Han2016}. Model compression-based methods involve iterative pruning and fine-tuning of networks and are often time-consuming processes.

On the other hand, knowledge distillation-based methods assume that the knowledge of a model is captured in its hidden layer activations and outputs. Hence, smaller models known as students receive supervision from larger models called teachers and the ground truths. In the case of classification, the probabilities assigned by the teacher to the incorrect classes constitute 'dark knowledge,' and it has been shown to improve the generalization ability of student models \cite{normal}. Based on the type of knowledge being transferred,  knowledge distillation methods fall into two broad categories: response-based and feature-based. In response-based methods, the knowledge is transferred by matching the outputs of the final layers of the teacher and the student models \cite{normal,ZSKD,FSKD,DAFL}, whereas in feature-based methods, the same is accomplished by matching the activations of the hidden layers of these models \cite{Romero2015,Heo2019,Zagoruyko2017}. Besides these broad categories, there exists another type of method for knowledge distillation, called self-KD. In this process, the teacher and the student models have the same capacities or are the same networks across different epochs of training \cite{selfKD_better_gen, self-kd1}. 

A crucial point that is often overlooked in the process is that different input data samples carry a different amount of information and a fixed loss function or target is not adequate to capture these sample-specific variations. The feature-based methods come closer to learning more sample-specific information than response-based methods. However, there is no proper explanation behind choosing the particular hidden layers from the teacher and student models to be involved in the knowledge transfer process, except for the matching output sizes. Also, supervision at the feature extraction level interferes with the student's feature extraction mechanism and prevents it from learning potentially better features than the teacher model. In addition to that, the existing methods are not efficient in terms of data utilization for knowledge distillation.

To overcome the above-mentioned problems, in this paper, sample-specific loss functions and targets based on the teacher's confidence are proposed for knowledge distillation. It is referred to here as Confidence Conditioned Knowledge Distillation (CCKD). If the teacher's predictions are correct, then more importance is assigned to its supervision than the ground truth and vice versa. CCKD incorporates this in two ways, namely through sample-specific loss functions (CCKD-L formulation) and sample-specific targets (CCKD-T formulation). The issue of data inefficiency is addressed by incorporating self-regulation into CCKD methods \cite{self-reg, self-reg2}. As a result, its sample efficiency improves a lot over other existing methods ~\cite{DAFL, ZSKD, meta, FSKD, normal}. It has been shown in the machine learning literature that a self-regulated learner that employs a metacognitive element to select what-to-learn and when-to-learn generalizes better than models that are trained in the conventional fashion \cite{meta3,meta1,meta2}. 

Experiments on benchmark datasets show that CCKD methods achieve better generalization than other state-of-the-art methods ~\cite{DAFL, ZSKD, meta, FSKD, normal} while being sample efficient in the process. Also, the student models trained by CCKD are less likely to repeat the same misclassifications as the teacher model on the training set. Experiments on the MNIST and the CIFAR10 datasets show that the student models can classify almost all the samples correctly that the teacher model misclassified during training. Following this finding, the student models were evaluated against adversarial attacks. Adversarial samples are created by adding noise to the existing samples in such a way that the resulting samples are misclassified by the model while still being visually insdistinguishable from the original samples. It is important to address adversarial attacks due to security reasons. It is an underexplored aspect in distillation research. Previous works in distillation literature explored single-pixel adversarial attacks \cite{defense_distill}, wherein pixels are altered one by one to the maximum allowable extent for crafting adversarial examples. In this work, robustness of the proposed methods to adversarial samples created through the Fast Gradient Sign Method is investigated \cite{advCreate}, wherein the entire image is perturbed at once. In addition to being better suited for knowledge transfer, experiments on benchmark datasets show that the CCKD methods also offer increased resistance to adversarial attacks than other distillation methods. Experiments show at least $3\%$ increase in performance against adversarial attacks for the MNIST and the Fashion MNIST datasets, and a $6\%$ increase for the CIFAR10 dataset. 

The contributions of this work can be summarized in the following points - 
\begin{itemize}
    \item Novel loss functions and targets are proposed that are tailored to accommodate different levels of information in different samples during the distillation process, based on the teacher model's confidence. These methods are named 'CCKD-L' and 'CCKD-T' based distillations respectively.
    \item Self-regulation is incorporated into the 'confidence conditioned target' based distillation (CCKD-T) to enhance its data efficiency. 
    \item It is shown that student models distilled through the proposed methods are less likely to repeat the same misclassifications as the teacher model on the original training set when compared to other distillation methods where the training set is available \cite{normal}.
    \item CCKD methods are evaluated on the MNIST, Fashion-MNIST, and CIFAR10 datasets and achieve similar or even better generalization performance as other state-of-the-art methods ~\cite{DAFL, ZSKD, meta, FSKD, normal} while being efficient in terms of data utilization. 
    \item Experiments show that CCKD methods offer increased resistance to adversarial attacks over other distillation methods. Student models distilled using CCKD methods are better able to resist adversarial attacks compared to student models distilled through other methods.
\end{itemize}

The rest of the paper is organized as follows - section 2 discusses related works, section 3 describes the methods proposed, section 4 describes the experiments and results and section 5 concludes the paper.

\section{Related Works}

The idea of knowledge distillation was proposed in \cite{org} and gained momentum in \cite{normal}. The student model is found to generalize better if supervised by the soft targets obtained at a high temperature from a bigger teacher model as opposed to the conventional way of training. This provides an easy method for transferring most of the generalization capacity of larger models to smaller models. Research in distillation is motivated by this observation. Apart from model compression, knowledge distillation has been successfully used in other applications as well. Recently, distillation has been applied in face recognition \cite{deepFace}, cross-modal hashing \cite{CMHash} and collaborative learning \cite{collaborative}. Several methods have been proposed for distillation and, a comprehensive review is provided in \cite{review} and \cite{review2}. Some of these methods are discussed in the subsections below.

\subsection{Response Based KD methods}
One of the earliest papers to appear in distillation literature was \cite{normal}. It proposed to use the original training data as the transfer set for distillation. In addition to learning from the ground truths, the student also receives supervision from the teacher model in the form of soft targets computed at a high softmax temperature. Subsequently, ~\cite{DAFL, ZSKD, meta} proposed methods for knowledge transfer wherein the transfer set was not available. The softmax space of the teacher network is modeled by using a Dirichlet distribution in \cite{ZSKD}. Synthetic data instances are then created by inverting the samples drawn from this distribution. The teacher model is used as a fixed discriminator in a generative adversarial framework in \cite{DAFL} to create synthetic data samples. Activation statistics of the teacher model are obtained during training and these are used to create synthetic data samples in \cite{meta}. Synthetic samples are created through feature inversion in \cite{ZSKD, meta}. These samples are then used for the knowledge distillation process. The process of distillation is the same as in ~\cite{normal}. A conditional distillation mechanism is proposed in ~\cite{conditional}. It incorporates the fact that the teacher model can sometimes be wrong in its predictions and only in that case the student must learn from the ground truths. 

\subsection{Feature Based KD methods}
The feature extraction process in the student model is guided by providing supervision from the hidden layers of the teacher model in \cite{Romero2015}. In case the output sizes of the layers involved in the transfer process do not match, a learnable convolutional regressor network is used to match the sizes. Transferring the activations of the hidden neurons instead of their response magnitudes is presented in \cite{Heo2019}. The authors show that generalization ability is better encoded by the decision boundaries formed by the hidden neurons rather than the actual response magnitudes. The loss function proposed in \cite{Heo2019} is sensitive to a neuron being active or inactive, so it is effective in transferring the decision boundaries of hidden neurons. A connector layer is used in case the intermediate layers of the teacher and student models engaged in distillation are of dissimilar sizes. Attention transfer is proposed as a mechanism for knowledge distillation in ~\cite{Zagoruyko2017}. Several types of attentions are defined by the authors for CNN layers. These attentions are computed at certain layers for the teacher and student models and are matched by minimizing the $L_p$ norm of their difference. The ground truths are also used to supervise the student models. 

\subsection{Self KD methods}
Self distillation presented in \cite{self-kd1} uses the same network as the teacher and student models. The model from the previous epoch is used as the teacher. Self distilltion can also be used as a tool for imposing regularization \cite{selfKD_better_gen} as it leads to better quality targets. The regularization properties of self distillation are explained in a rigorous mathematical framework in \cite{self_distill_hilbert}. Class-wise consistent predictions resulting from self distillation are explored in \cite{selfKD_reg}. Self distillation is used as a mode of regularization to minimize intra class variations in predictions. Predictions on different samples belonging to the same class are distilled to mitigate overconfident predictions. Self distillation has also been used in natural language processing \cite{selfKD_nlp}. Supervising the same model at different depths is explored in \cite{ownTeacher}. It falls into self distillation methods as the teacher and the student models are essentially the same. As a result, deployment can be flexible as the user can trade off accuracy for execution speed and vice versa.

The proposed work is different from self distillation methods as the teacher and student models are different networks of different capacities. Hence, self distillation methods are not used as baselines for comparison. The proposed CCKD-T method is similar in nature to \cite{selfKD_better_gen} but the student and teacher models used are of different sizes. Also, the proposed framework is more general and is aimed at improving distillation performance. Unlike \cite{selfKD_better_gen}, the proposed framework is not aimed at providing additional regularization. Experiments are designed to specifically establish the superiority of the proposed methods in improving the performance of student models for minimizing the repetition of teacher's misclassifications and resisting adversarial attacks. Unlike \cite{selfKD_better_gen}, the proposed methods address transferring different levels of knowledge present in different samples which leads to better performing student models. Data efficiency is also a concern which the proposed methods address by incorporating self-regulation (section 2.5) but it is absent in \cite{selfKD_better_gen}.

\subsection{KD and Adversarial Attacks}
The first work to propose the use of self distillation to resist adversarial attacks was \cite{defense_distill}. It is shown that self distillation reduces the gradient amplitudes responsible for the creation of adversarial samples as it makes the models more smooth. As a result, the model's sensitivity to adversarial perturbations decreases significantly. In addition to that, it is shown that self distillation also improves the robustness of the models. That is, the average number of features that need to be perturbed to produce adversarial samples increases significantly after self distillation. The type of attack investigated in the study was modifying a few pixels by a large amount. This helps in quantifying the number of features changed in crafting adversarial samples. 

Distillation aims at improving the generalization performance of student models. One way of doing this is by making the decision boundaries of the teacher and student models as similar as possible. Concepts from adversarial sample creation have been successfully leveraged to improve distillation performance. For example, an adversarial generator network is trained in \cite{advBelief} to find regions in the input space where there is a high mismatch between the teacher and the student's predictions. Inputs were then sampled from these regions for the distillation process and it was shown that such a method for knowledge transfer leads to better agreement between the teacher and student classifiers' decision boundaries. The framework proposed in \cite{advBelief} is a data-free distillation framework. The same idea as above is used in \cite{advDecision} to craft samples that are close to the teacher model's decision boundary and then use these for knowledge transfer. The authors postulate that the samples that are closer to the decision boundary are more effective in the knowledge transfer process. 

\subsection{Self Regulation}
In the conventional deep neural network training and knowledge distillation methods, all the training samples participate equally in capturing the input-output relationship. However, in machine learning literature, it has been shown that self-regulation helps in better generalization process \cite{meta3,meta1,meta2}. These meta-cognitive neural networks (originally proposed in the context of human learning in \cite{metamem}) employ self-regulation to select appropriate training samples to learning from stream-of-training data. The heuristic strategy helps in identifying what-to-learn, when-to-learn, and how-to-learn. Thus, the teacher network needs to employ self-regulation to identify the potential samples required for acquiring the knowledge and train the student network based on selected samples and their sample significance values. 

Self-regulation is incorporated into distillation for the first time in \cite{self-reg2}. The authors show that all of the samples in the training data are not equally important towards effective knowledge transfer. Student models can achieve similar levels of generalization as the teacher models by using significantly lesser number of samples while employing self-regulation for sample selection. Self-regulation succesfully enhances the data efficiency of conventional distillation.

\section{Confidence Conditioned Knowledge Distillation Methods}
In this section, the proposed Confidence Conditioned Knowledge Distillation (CCKD) methods are described in detail. First, we briefly summarize knowledge distillation and then present the CCKD methods - Confidence Conditioned Loss Based KD (CCKD-L), Confidence Conditioned Target Based KD (CCKD-T), and Confidence Conditioned Target Based KD with self-regulation CCKD-T + self-reg). 

\begin{figure*}[t]
    \centering
    \includegraphics[width=\textwidth]{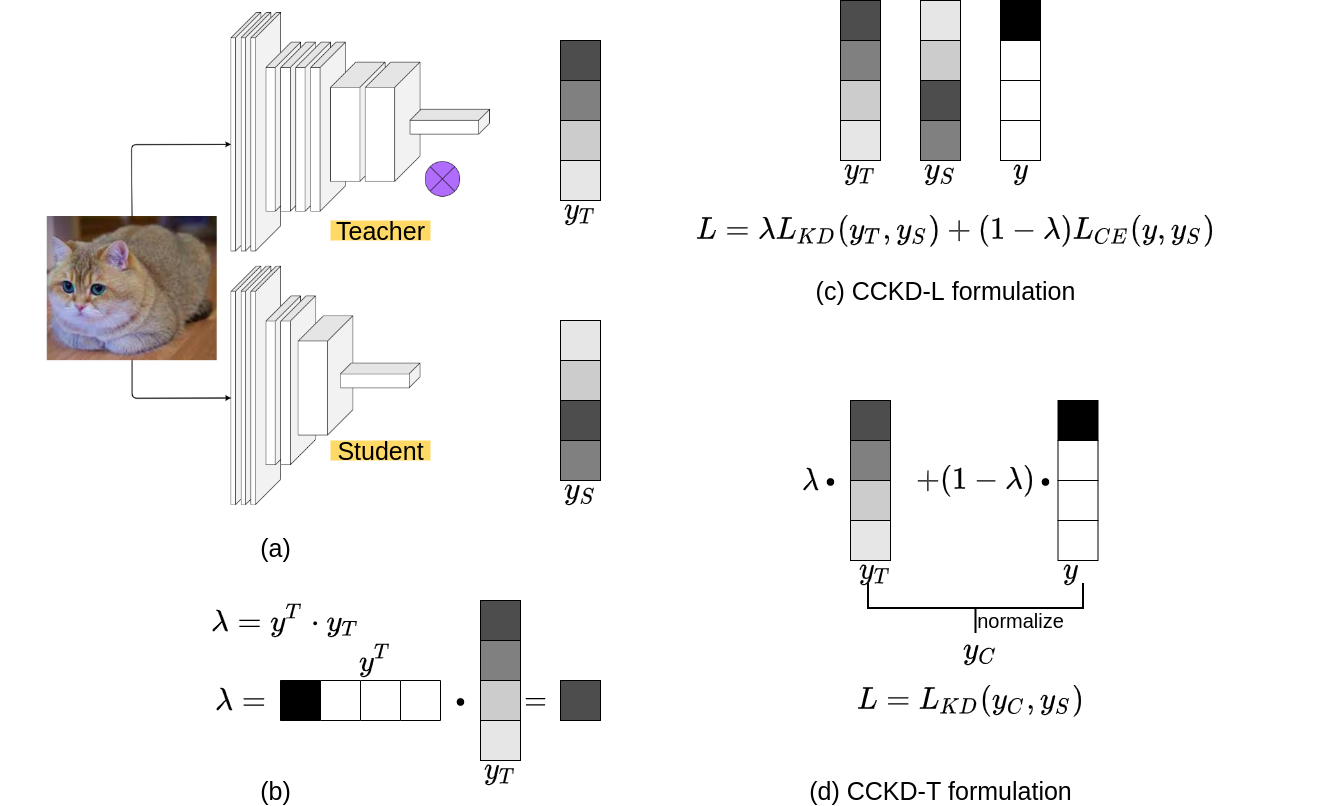}
    \caption{Schematic diagram illustrating the idea behind CCKD-L and CCKD-T.  (a) Predictions of the teacher and the student models. The little blue cross near the teacher model depicts that it is pretrained and its parameters are not updated during distillation. (b) Computation of the confidence assigned by the teacher to the correct class $\lambda = y^T\cdot y_T$. (c) In CCKD-L (CC loss) formulation, the losses are weighted according to the teacher's confidence. (d) In CCKD-T (CC target) formulation, the same thing is done with the labels. $L_{CE}$ is the cross entropy loss and $L_{KD}$ is the distillation loss.}
    \label{fig:Figure 1}
\end{figure*}

\subsection{Knowledge Distillation}
In conventional knowledge distillation, the soft targets computed from the teacher model at a temperature $\tau$ are used. Given a teacher network $T$ parametrized by $\theta_T$ and a student network $S$ parametrized by $\theta_S$, distillation minimizes the following objective over all samples $(x, y)$ in the transfer set $\mathbb{D}$:

\begin{equation}
L = \sum_{\substack{(x, y) \in \mathbb{D}}} L_{KD}(S(x, \theta_S, \tau), T(x, \theta_T, \tau)) + \lambda L_{CE}(\hat{y}_S, y)
\end{equation}

where, $L_{KD}$ is the distillation loss which is minimized at a temperature $\tau$. It can be the cross entropy loss for classification or the $L_2$ loss for regression. $L_{CE}$ is the cross entropy loss which is minimized at a temperature of 1. $\hat{y}_S$ is the prediction of the student network on the sample $x$ at temperature 1, and $\lambda$ is a hyperparameter to balance the two losses. It is important to note that the conventional KD process assigns a fixed relative importance $\lambda$ to the loss incurred on the ground truths $y$.

\subsection{Confidence Conditioned Loss based KD (CC Loss or CCKD-L)}
%
%

The loss function is modified to reflect the confidence of the teacher model in its predictions, thereby accounting for the possibility that the teacher could be wrong in its predictions. So unlike the conventional KD case, this formulation does not have a fixed loss function. Instead it adapts to the different levels of information to be distilled from different samples. The loss function is sample specific. It is given by:

\begin{equation}
    L = \lambda L_{KD}(y_T, y_S) + (1 - \lambda) L_{CE}(y, \hat{y}_S) 
\end{equation}

where, $y_T, \ y_S$ in $L_{KD}$ are the predictions of the teacher and the student models respectively on a sample $x$ computed at a softmax temperature of $\tau > 1$. $\hat{y}_S$ in $L_{CE}$ is the prediction of the student model on the same sample $x$ at a softmax temperature of $1$. $\lambda$ is the confidence assigned by the teacher to the correct class. Mathematically, it can be computed as an inner product between $y_T$ and $y$: 

\begin{equation}
    \lambda = y^Ty_T
\end{equation}

So if the teacher is highly confident in predicting the correct class, more importance is assigned to the teacher's supervision and vice versa. It is important to note that $\lambda$ varies across samples, but for a particular sample, it always stays the same because the teacher model is pre-trained.

\subsection{Confidence Conditioned Target based KD (CC Target or CCKD-T)}
Before moving onto confidence conditioned targets, another variant of distillation called teacher only distillation is introduced. Teacher only distillation uses $y_T$ computed at a softmax temperature of $\tau$ with $\lambda = 0$ (equation 1). Separate training uses only $y$ and $\hat{y}_S$ used in $L_{CE}$ is computed at a temperature of 1. So, for separate training, $L_{KD}$ is not used and $\lambda = 1$ (equation 1). Separate training therefore refers to the conventional way of training models whereas teacher only distillation relies exclusively on supervision from the teacher model. These will be used as the baselines in the experiments on the repetition of teacher model's mistakes by the student model and the resistance of student models to adversarial attacks.

Confidence conditioned targets is a reformulation of the confidence conditioned loss objective in the output space. Confidence conditioned targets is a soft intermediate between teacher only distillation and separate training of the student models. Confidence conditioned targets are mathematically expressed as:

\begin{equation}
    \bar{y} = \lambda y_T + (1 - \lambda)y
\end{equation}

\begin{equation}
    y_C = \frac{\bar{y}}{||\bar{y}||_1}
\end{equation}

The $L_1$ normalization is important to ensure that the confidence conditioned targets $y_C$ is a valid probability distribution. It is important to note that even though the targets are fixed because the teacher model is pre-trained, they include sample specific information in the form of the confidence of the teacher model $\lambda$. The loss function to be used is then given by:

\begin{equation}
    L = L_{KD}(y_C, y_S)
\end{equation}

where $y_S$ are the predictions from the student model computed at a softmax temperature of $\tau$. A schematic representation of the methods described in sections 3.2 and 3.3 is given in Figure 1.


\subsection{Confidence Conditioned Knowledge Distillation on Target with Self Regulation (CCKD-T+Reg)}

The next step is to improve the data utilization efficiency of distillation. This is done by introducing self regulation \cite{self-reg}. As pointed out in \cite{meta3,meta1,meta2}, all the samples in the input data are not required for learning the input output relationship. The proposed implementation of CCKD-T+Reg closely follows the formulation of self-regulation for distillation in \cite{self-reg2}. While employing self regulation, the model need not learn on a sample again if it is already too confident on it. So the model is able to distinguish between easy and hard samples based on an epoch dependent threshold and discards the easy samples from the process. This introduces additional regularization and is expected to help the models generalize better. The self regulation process is explained below and it is added to the confidence conditioned targets formulation.

Given a dataset $\mathbb{D}$ containing labeled samples $(x, y)$ and a student model $S$, the following quantities are monitored for all samples in all epochs $(N)$:

\begin{itemize}
    \item The predicted label, $\hat{y}$: \\$\hat{y} = \argmax y_S = \argmax S(x)$
    
    \item The difference between the maximum and the second maximum predicted probabilities, $\delta$: \\$\delta = \max S(x) - \max \{m | m \in S(x), \ m \neq \max S(x)\}$
\end{itemize}

As the model learns to classify properly, the difference $\delta$ gradually increases with the number of epochs $n$. A sample is included in training if the predicted class is incorrect or if $\delta$ is less than an epoch dependent adaptive threshold, $\eta$. $\delta$ will increase faster for easy samples compared to difficult samples. The purpose of the epoch dependent threshold function $f(n)$ is to filter out such samples from training. Since $\delta$ is the difference between the maximum and the second maximum probabilities, it is in the range $[0, 1]$. So the function $f(n): \mathbb{N} \rightarrow [0, 1]$ must be an increasing function of $n$. So $f(n) = 1-\exp(-\alpha n)$ is chosen as threshold predictor, where $\alpha > 0$ is a hyperparameter. It maximizes the difference in the predicted posterior probabilities by allowing samples with a smaller growth rate of $\delta$ to participate more in training. The method is shown in Figure 2. The algorithm for including self-regulation with confidence conditioned targets is shown in Algorithm 1. This is the self-regulation process explained briefly, and it is applied to the student model during distillation. The teacher model is trained in the conventional manner, and its parameters are not updated during the process. 

\begin{figure}[h]
    \centering
    \includegraphics[width=\linewidth]{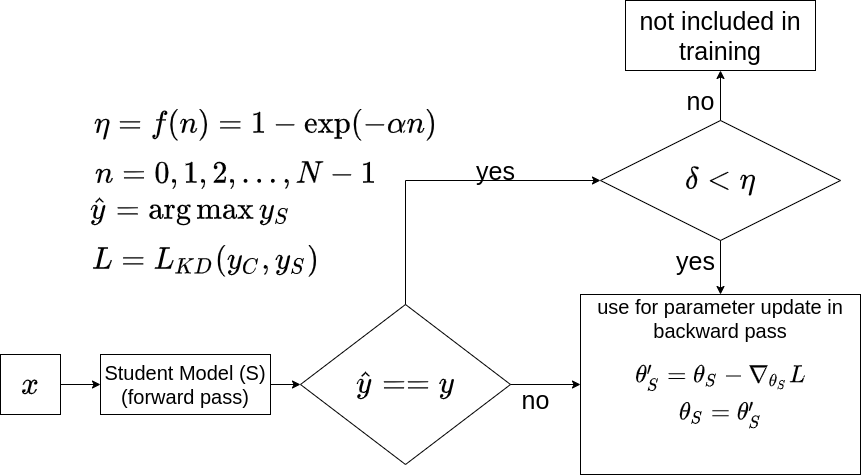}
    \caption{Self Regulation Algorithm: A sample $(x, y)$ is included in training if it is predicted incorrectly by the student model, or if the difference between the maximum and the second maximum predicted probabilities ($\delta$) is less than an epoch dependent threshold ($\eta$). The targets for distilation $y_C$ are the CCKD-T targets computed using equations 4 and 5. $y_S = S(x)$ is the prediction of the student model $S$ on the sample $(x, y)$.}
    \label{fig:Figure 2}
\end{figure}

\begin{algorithm}[!htbp]
\textbf{Input} Pre-trained Teacher network $T$, Student network $S$ with parameters $\theta_S$ (without output softmax), dataset $\mathbb{D} = \{(x_i, y_i)\}_{i=1}^N$, epochs $n$, parameter $\alpha$ for self-regulation, temperature $\tau$ for distillation, learning rate $\gamma$\\
\textbf{Output} Parameters of trained Student network $\theta_S$\\
\begin{algorithmic}[1]
\FOR{$i$ in range($n$):}
\STATE $\eta = 1 - \exp{(\alpha i)}$
\FOR{$(x, y) \in \mathbb{D}$:}
\STATE $z_T, \ z_S = T(x), \ S(x)$
\STATE $y_T, \ y_S = softmax(z_T/\tau), \ softmax(z_S/\tau)$
\STATE $\hat{y} = \argmax y_S$
\STATE compute $\delta$ from $y_S$ 
\STATE $\lambda = y^Ty_T$
\IF{$\hat{y} \neq y$ or ($\hat{y} == y$ and $\delta < \eta$):}
\STATE compute $y_C$ from equations (4), (5)
\STATE compute $L$ from equation (6)
\STATE update student's parameters: $\theta_S' = \theta_S - \gamma\nabla_{\theta_S} L$
\STATE $\theta_S = \theta_S'$
\ELSE
\STATE continue
\ENDIF
\ENDFOR
\ENDFOR
\RETURN $\theta_S$
\end{algorithmic}
\caption{Confidence Conditioned Knowledge Distillation on Target with Self Regulation (CCKD-T+Reg)}
\end{algorithm}

\section{Experiments and Results}
In this section, the CCKD methods are evaluated for the following propoerties:
\begin{itemize}
\item Generalization Performance.
\item Sample Efficiency.
\item Repetition of the misclassifications of the teacher by the student. 
\item Resistance to adversarial attacks.
\end{itemize}
 
 The MNIST, Fashion MNIST, SVHN and CIFAR10 datasets are used for experiments. The following sections describe the training and evaluation protocols and results.

\subsection{Datasets and Metrics}
\begin{itemize}
    \item \textbf{MNIST} It is a dataset of handwritten digits. Each sample is a 28$\times$ 28 grayscale image. The training set has 60000 labeled samples and the test set has 10000 labeled samples. Lenet-5 model is used as the teacher network and Lenet-5 half model as the student network. 
    
    \item \textbf{Fashion-MNIST (FMNIST)} It is a dataset of 10 fashion items. Each sample is a 28$\times$ 28 grayscale image. The training set has 60000 labeled samples and the test set has 10000 labeled samples. Lenet-5 model is used as the teacher network and Lenet-5 half model as the student network.
    
    \item \textbf{SVHN} It is a dataset of housing numbers. It contains the same digits as there in the MNIST dataset, with the exception that the digit 0 is labeled 10. The SVHN dataset is more challenging and realistic than the MNIST dataset. It contains 73257 digits for training and 26032 digits for testing. Each sample is a 32 $\times$ 32 colour image. Alexnet model is used as the teacher network and Alexnet half model as the student network in one setting. In the other setting, ResNet 34 \cite{He2016} model is used as the teacher network and the ResNet 18 model as the student network.
    
    \item \textbf{CIFAR10} It is a dataset of 10 items as classes and each class contains 6000 samples. Each sample is a 32 $\times$ 32 colour image. The training set contains 50000 labeled samples and the test set contains 10000 labeled samples. Alexnet model is used as the teacher network and Alexnet half model as the student network 
\end{itemize}

The batch size is set to 512, the distillation temperature $\tau$ is set to 20, the normal temperature is set to 1, and the hyperparameter $\lambda$ in normal distillation is set to 0.3 \cite{ZSKD}. The models are trained using the Adam optimizer \cite{adam}. Models are evaluated at the normal temperature using the accuracy on the test set. To ensure a fair comparison, the learning rate is kept the same across all training and distillation methods. Across all datasets, a learning rate of 0.001 is used for training the teacher and student models separately whereas during distillation, a learning rate of 0.01 is used. The epochs are also kept the same across all methods including separate training of teacher and student models. The details of epochs and network sizes are given in Table 1. Two NVIDIA GeForce RTX 2080 Ti cards are used for the experiments. Since the proposed CCKD methods are response based, other response based methods available in the literature ~\cite{DAFL, ZSKD, meta, FSKD, normal} are chosen as baselines for comparison. All implementations are done in pytorch \cite{Paszke2017}.
%

 \begin{table}[!htbp]
 \centering
\begin{tabular}{clllll}
\hline
\multicolumn{1}{l}{\multirow{2}{*}{}} & \multicolumn{2}{c}{\textbf{Model}}                                          & \multicolumn{2}{c}{\textbf{Parameters}}                                     & \multicolumn{1}{c}{\multirow{2}{*}{\textbf{Epochs}}} \\ \cline{2-5}
\multicolumn{1}{l}{}                  & \multicolumn{1}{c}{\textbf{Teacher}} & \multicolumn{1}{c}{\textbf{Student}} & \multicolumn{1}{c}{\textbf{Teacher}} & \multicolumn{1}{c}{\textbf{Student}} & \multicolumn{1}{c}{}                                 \\ \hline
\textbf{MNIST}                        & LeNet5                               & LeNet5 Half                          & $\sim$62K                            & $\sim$36K                            & 200                                                  \\ \hline
\textbf{FMNIST}                       & LeNet5                               & LeNet5 Half                          & $\sim$62K                            & $\sim$36K                            & 200                                                  \\ \hline
\textbf{CIFAR10}     & AlexNet                              & AlexNet Half                         & $\sim$1.66M                          & $\sim$0.4M                           & 1000                                                 \\
\hline
\multirow{2}{*}{\textbf{SVHN}}        & AlexNet                              & AlexNet Half                         & $\sim$1.66M                          & $\sim$0.4M                           & 500                                                  \\
                                      & ResNet34                             & ResNet18                             & $\sim$21M                            & $\sim$11M                            & 200                                                  \\ \hline
\end{tabular}
 \label{Table:1}
 \caption{Network Architectures and epochs.}
 \end{table}


\subsection{Distillation Results on MNIST}
The teacher model trained separately achieves an accuracy of 0.9914, and the student model trained separately achieves an accuracy of 0.9900. The performance of the proposed methods and other available state-of-the-art methods on the MNIST dataset are reported in Table 2. The proposed methods perform at least as better as the other state-of-the-art methods. Since the MNIST dataset is comparatively simple, all methods obtain such high accuracy values leaving very little scope for improvement.

\begin{table}[]
\centering
\begin{tabular}{ll}
\hline
\multicolumn{1}{c}{\textbf{Method}} & \multicolumn{1}{c}{\textbf{Accuracy}} \\ \hline
Normal   \cite{normal}                          & 0.9925                                \\
Few Shot KD       \cite{FSKD}                  & 0.8670                                \\
Meta Data         \cite{meta}                  & 0.9247                                \\
Zero Shot KD      \cite{ZSKD}                  & 0.9877                                \\
Data Free KD      \cite{DAFL}                  & 0.9820                                \\
T + KEGNET        \cite{no_data}                  & 0.9632                                \\ \hline
\multicolumn{2}{l}{\textbf{Proposed Methods}}                                           \\ \hline
CCKD-L                              & 0.9909                                \\
CCKD-T                              & 0.9896                                \\
CCKD-T + Reg                        & 0.9878                                \\ \hline
\end{tabular}
\label{Table:2}
\caption{Comparison of Test Set Accuracy on MNIST Dataset.}
\end{table}

\subsection{Distillation Results on Fashion MNIST}
The teacher model trained separately achieves an accuracy of 0.9004, and the student model trained separately achieves an accuracy of 0.8939. The performance of the proposed methods and other available state-of-the-art methods on the Fashion MNIST dataset are reported in Table 3. CCKD-T method obtains the best performance amongst all the proposed methods, showing an improvement of more than 15\% over \cite{FSKD} and nearly 10\% over \cite{ZSKD}. It comes very close to \cite{normal}. The same can be said about the CCKD-L method as well. The CCKD-T+Reg method comes really close to \cite{no_data}.

\begin{table}[]
\centering
\begin{tabular}{ll}
\hline
\multicolumn{1}{c}{\textbf{Method}} & \multicolumn{1}{c}{\textbf{Accuracy}} \\ \hline
Normal       \cite{normal}                       & 0.8966                                \\
Few Shot KD        \cite{FSKD}                 & 0.7250                                 \\
Zero Shot KD       \cite{ZSKD}                & 0.7962                                                               
\\
T + KEGNET        \cite{no_data}                  & 0.8780                                \\ \hline
\multicolumn{2}{l}{\textbf{Proposed Methods}}                                           \\ \hline
CCKD-L                              & 0.8888                                \\
CCKD-T                              & 0.8900                                \\
CCKD-T + Reg                        & 0.8773                                \\ \hline
\end{tabular}
\label{Table:3}
\caption{Comparison of Test Set Accuracy on Fashion MNIST Dataset.}
\end{table} 

\subsection{Distillation Results on CIFAR10}
The teacher network trained separately achieves an accuracy of 0.8268, and the student model trained separately achieves an accuracy of 0.8303. The performance of the proposed methods and other available state-of-the-art methods on the CIFAR10 dataset are reported in Table 4. CCKD-T method performs the best amongst all the proposed methods, obtaining nearly 10\% improvement over \cite{ZSKD} and coming really close to \cite{normal}. The same thing can be said about the CCKD-L and the CCKD-T+Reg methods as well.


\begin{table}[]
\centering
\begin{tabular}{ll}
\hline
\multicolumn{1}{c}{\textbf{Method}} & \multicolumn{1}{c}{\textbf{AlexNet}}\\ \hline
Normal    \cite{normal}                          & 0.8008                                                                                               \\                                
Zero Shot KD    \cite{ZSKD}                    & 0.6956                                                                \\
\hline
\multicolumn{2}{l}{\textbf{Proposed Methods}}                                                                                \\ \hline
CCKD-L                              & 0.7930                                                             \\
CCKD-T                              & 0.7979                                                             \\
CCKD-T + Reg                        & 0.7787                                                             \\ \hline
\end{tabular}
\label{Table:4}
\caption{Comparison of Test Set Accuracy on the CIFAR10 Dataset.}
\end{table}

\subsection{Distillation Results on SVHN}
With the AlexNet models, the teacher network trained separately achieves an accuracy of 0.9380, and the student model trained separately achieves an accuracy of 0.9326. With the ResNet models, the teacher network trained separately achieves an accuracy of 0.9464, and the student model trained separately achieves an accuracy of 0.9445. The performance of the proposed methods and other available state-of-the-art methods on the SVHN dataset are reported in Table 5. The proposed methods perform significantly better than \cite{no_data}, showing an improvement of at least 6\% in each case.

\begin{table}[]
\centering
\begin{tabular}{lll}
\hline
\multicolumn{1}{c}{\textbf{Method}} & \multicolumn{1}{c}{\textbf{AlexNet}} & \multicolumn{1}{c}{\textbf{ResNet}} \\ \hline
T + KEGNET \cite{no_data}                          & N/A                                  & 0.8726                              \\ \hline
\multicolumn{3}{l}{\textbf{Proposed Methods}}                                                                                \\ \hline
CCKD-L                              & 0.9219                               & 0.9522                              \\
CCKD-T                              & 0.9247                               & 0.9539                              \\
CCKD-T + Reg                        & 0.9268                               & 0.9531                              \\ \hline
\end{tabular}
\label{Table:5}
\caption{Comparison of Test Set Accuracy on the SVHN Dataset.}
\end{table}

FashionMNIST, SVHN and CIFAR10 are more realistic datasets compared to MNIST, so the performance does not reach beyond 90\%. The previous comparison gives an assurance that CCKD methods are at least as good as the baselines while using full data. It is observed that CCKD-T method performs slightly better than the other methods (CCKD-L and CCKD-T+Reg) in general. 

\subsection{Sample Efficiency on adding Self-Regulation}
In this section, the sample efficiency of the method proposed in section 3.3 (CCKD-T+Reg) is evaluated. The results are tabulated in Table 6. The value of $\alpha$ used in the threshold function for self-regulation is also provided. This is evaluated for the MNIST, Fashion MNIST and CIFAR10 datasets.

In the MNIST dataset, there are 60000 training samples and the distillation takes place over 200 epochs. So the student model sees a total of $60000\times 200 = 12000000$ samples during normal \cite{normal} and teacher only distillations. This is expected as these methods use the full data. In comparison to these methods, CCKD-T+Reg method uses fewer samples due to self-regulation. Adding self-regulation decreases the amount of data required to achieve a comparable level of generalization performance. In general, a slight decrease in performance is observed (Tables 2-5) as CCKD-T+Reg method does not use all the samples present in the dataset across all epochs. For the CIFAR10 dataset, the sample efficiency results for the AlexNet case are reported. CIFAR10 dataset is more realistic compared to MNIST and Fashion-MNIST datasets, so the sample utilization is the highest. 


\begin{table}[]
\centering
\begin{tabular}{ccc}
\hline
\textbf{Dataset} & \textbf{Sample Efficiency}    & \multicolumn{1}{c}{$\mathbf{\alpha}$} \\ \hline
MNIST            & 103476/12000000 (0.8623\%)    & 0.01                                                \\
FMNIST           & 1880714/12000000 (15.6726\%)  & 0.01                                                \\
CIFAR10          & 33101348/50000000 (66.2027\%) & 0.008                                               \\ \hline
\end{tabular}
\label{Table:6}
\caption{Sample Efficiency on adding self-regulation.}
\end{table}

\subsection{Do the students repeat the same misclassifications as the teacher?}
To establish the beneficial properties of the CCKD methods for knowledge transfer, their ability to transfer the misclassifications of the teacher model is investigated. Specifically, how much of the teacher's misclassifications do the student models retain? and, at what rate do the student models commit new mistakes? A method better suited for knowledge transfer is expected to have low values for both these quantities. This section evaluates the misclassifications of the students trained by CCKD methods with respect to their teachers. For this purpose, the normal distillation \cite{normal} and the teacher only distillations are used as baselines. Teacher only distillation does not use ground truths for training the student as explained in section 3.3. The evaluation is carried out on the MNIST and the CIFAR10 datasets.

Two quantities are defined for evaluating this - the success and failure rates. Mathematically, the formulation is as follows. Consider a training dataset $\mathbb{D}$, a teacher model $T$ and a student model $S$. Let $C_T$ and $W_T$ denote the subsets of $\mathbb{D}$ that are classified correctly and wrongly by the teacher respectively. Similarly, let the corresponding subsets for the student model be denoted by $C_S$ and $W_S$ respectively. 
The success rate $\eta_S$ is given by:
\begin{equation}
    \eta_S = \frac{|W_T \cap C_S|}{|W_T|}
\end{equation}
And, the failure rate $\eta_F$ is given by:
\begin{equation}
    \eta_F = \frac{|C_T \cap W_S|}{|C_T|}
\end{equation}

 In other words, success rate is a measure of the method's resistance to transmit the misclassifications of the teacher model to the student model during distillation. Failure rate, on the other hand is a measure of the new mistakes that the student model commits in comparison to the teacher model. It is evident that failure rate will be positive because the student model is of lesser capacity than the teacher model. From the above definitions, a higher success rate and a lower failure rate is a clear indicator of a better method for knowledge transfer. 
 
\begin{table}[!htb]
\begin{tabular}{c|c|cc|cc}
\hline
\multicolumn{2}{l|}{\multirow{2}{*}{}}                                    & \multicolumn{2}{c|}{\textbf{MNIST}}       & \multicolumn{2}{c}{\textbf{CIFAR10}}      \\ \cline{3-6} 
\multicolumn{2}{l|}{}             & $\eta_S$ & $\eta_F$              & $\eta_S$ & $\eta_F$               \\ \hline
\multirow{2}{*}{\textbf{Baselines}}        & Normal \cite{normal} & 0.035    & $4.2 \times 10^{-4}$  & 0.127    & $6.86 \times 10^{-4}$  \\
                                  & Teacher Only                          & 0.035    & $7.2 \times 10^{-4}$  & 0.129    & $1.2 \times 10^{-3}$   \\ \hline
\multirow{3}{*}{\textbf{Proposed Methods}} & CCKD-L                                & 1.0      & $1.67 \times 10^{-5}$ & 0.995    & $2.018 \times 10^{-5}$ \\
                                  & CCKD-T                                & 1.0      & $1.67 \times 10^{-5}$ & 0.995    & $2.018 \times 10^{-5}$ \\
                                  & CCKD-T + Reg.                         & 1.0      & $1.67 \times 10^{-5}$ & 0.995    & $2.018 \times 10^{-5}$ \\ \hline
\end{tabular}
\label{Table:7}
\caption{Success and Failure Rates for MNIST, and CIFAR10 datasets.}
\end{table}
 
 Table 7 shows that the baseline methods have low values for both the quantities. This shows that the baseline methods transfer most of the mistakes of the teacher model to the student model, even though the student model does not commit many new mistakes. Thus, the baseline methods transfer both - the misclassifications as well as the correct classifications of the teacher model to the student model. Table 7 also shows that CCKD methods have higher success rates and lower failure rates compared to the baseline methods establishing that these are better suited for knowledge transfer. It can be expected that models with better generalization ability have higher success rates and lower failure rates compared to methods with lower generalization ability. Since other state-of-the-art methods presented in Tables 2-5 have lower levels of generalization than the normal distillation method \cite{normal} baseline, the CCKD methods are expected to be better than these methods as well. CCKD methods perform much better in terms of transferring these misclassifications by accounting for the different levels of information present in different samples through sample specific losses and targets. This is a major benefit of CCKD methods. The comparison in Table 7 shows that confidence conditioned targets are better suited for knowledge transfer than the soft labels prevalent in response based distillation. However, the student model makes misclassifications on other samples as it is of lesser capacity than the teacher model.

\subsection{Effectiveness Against Adversarial Attacks}
Following the previous experiments that reveal the beneficial properties of the proposed CCKD methods for knowledge transfer, these are evaluated against adversarial samples. Since student models trained with these methods do not retain most the misclassifications of the teacher model, it is expected that their performance on adversarial samples will also be better than students trained with other distillation methods. Adversarial samples are obtained by slightly perturbing the input data samples for the model. This perturbation results in a misclassification, but the change is too subtle to be perceptually significant. The original and the perturbed samples will appear indistinguishable to the human eye. The existence of such samples pose a threat against security and privacy as such samples can be succesfully leveraged by malevolent attackers to extract valuable information and evade rules. It is important to address them, especially in the context of distillation, as the student model is often of a lesser capacity than the teacher model. \cite{defense_distill} was the first work in distillation literature to address this. It evaluated the robustness of self distillation against single-pixel adversarial attacks. In single-pixel attacks, the image is altered one pixel at a time, applying a maximum allowable perturbation to each pixel that is changed. This gives an explicit measure of the number of features that need to be altered to execute a successful attack. \cite{defense_distill} used the MNIST dataset for their experiments. MNIST dataset contains grayscale images of digits. The background pixels are more in number than the pixels constituting the actual digits. Single-pixel and structured attacks are suitable in such a simplified setting. For example, the pixels near the top end of the digit 1 can be modified so that it looks more like the digit 7. In more practical cases, for example, like the CIFAR10 dataset, which contain colored images of objects, it is not straight forward to execute such attacks. Hence, the entire image needs to be perturbed at once. The strength of noise required to execute a successful attack is a measure of the robustness and generalization ability of the model. So, the proposed methods are evaluated against adversarial samples created by the Fast Gradient Sign Method \citep{advCreate}. Next, the process of adversarial sample creation is described briefly followed by the experiments and results.

The student network that is trained separately (without supervision from any teacher model) is used in the adversarial sample creation process. Let $f$ denote the student model. Let $(x, y)$ be a sample from the training or testing set and $x_A$ be the adversarial sample crafted from it. Then $x_A$ is obtained by iteratively applying gradient ascent through the following equation:

\begin{equation}
x_A = x + \epsilon \text{sign} (\nabla_x J(y, f(x)))
\end{equation}

where $J(y, .)$ denotes the cross entropy loss function which was used to train the student $f$. $\epsilon$ is a user specified value of the step size. This also controls the strength of the perturbation that is applied.

Adversarial examples are crafted from the examples present in the training set for the MNIST, Fashion MNIST and CIFAR10 datasets. The student network, the value of $\epsilon$ and the number of samples crafted for each dataset case are tabulated in Table 8. Figures 3 - 5 show some adversarial samples crafted on these datasets. The first number in the heading is the true class id, and the second number is the class id which the model predicts after adversarial perturbation is applied.

\begin{figure}
\centering
    \includegraphics[width=0.5\linewidth]{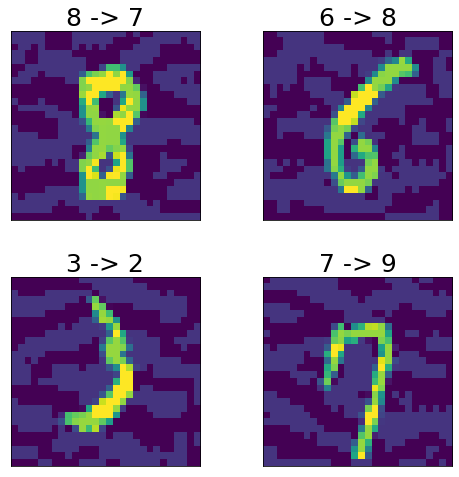}
    \label{Figure: 3}
    \caption{Adversarial examples on MNIST}
\end{figure}

\begin{figure}
\centering
    \includegraphics[width=0.5\linewidth]{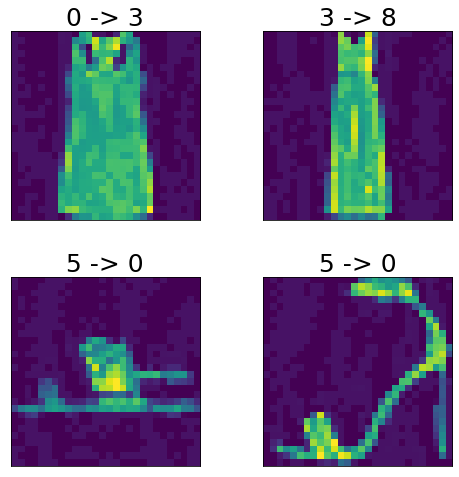}
    \label{Figure: 4}
    \caption{Adversarial examples on FashionMNIST}
\end{figure}

\begin{figure}
\centering
    \includegraphics[width=0.5\linewidth]{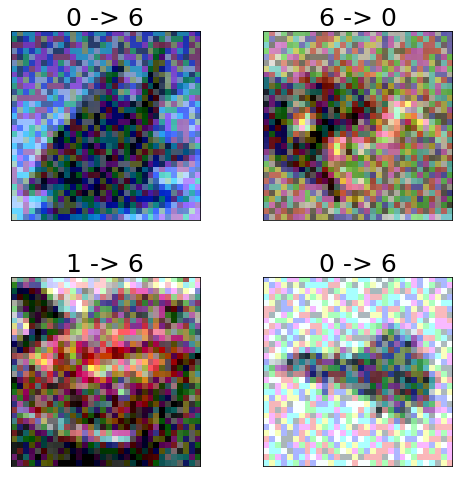}
    \label{Figure: 5}
    \caption{Adversarial examples on CIFAR10}
\end{figure}


\begin{table}[!htb]
\centering
\begin{tabular}{lllll}
\hline
                           & \multicolumn{1}{c}{\textbf{MNIST}} & \multicolumn{1}{c}{\textbf{FMNIST}} & \multicolumn{1}{c}{\textbf{CIFAR10}}  \\ \hline
\textbf{Model}             & LeNet5 Half                            & LeNet5                              Half & AlexNet Half                                                        \\
\textbf{step size(s)}      & 0.15                               & 0.05                                & 0.15                                                         \\
\textbf{number of samples} & 30000                              & 30000                               & 2500                                                             \\ \hline
\end{tabular}
\label{Table:8}
\caption{Details of adversarial sample creation.}
\end{table}

Table 9 shows the effectiveness of the proposed methods against adversarial samples. The student models distilled using the proposed methods and the normal \cite{normal} and teacher only methods are evaluated on the adversarial samples crafted earlier, and the accuracy of classification is reported. This is an indicator of the model's ability to resist adversarial attacks.

Since the teacher model has higher capacity than the student model, it is expected that the student models trained through distillation will demonstrate higher resistance to adversarial attacks in general. This is reflected by the generally high values in the table. This shows that due to the high capacity of the teacher model, it is able to classify most of the adversarial examples correctly and this ability is transferred to the student through distillation. As observed in all cases, the CCKD methods are better than the normal \cite{normal} and the teacher only methods in resisting adversarial attacks. Thus the CCKD methods are better at transferring the robustness of the teacher model to the student model. In general, the CCKD-T method with self regulation is expected to give better resistance because self regulation introduces additional regularization (section 3.4). 

\begin{table}[!htb]
\centering
\begin{tabular}{cccccl}
\hline
\multicolumn{2}{l}{}                                  & \textbf{MNIST}                   & \textbf{FMNIST}                  & \textbf{CIFAR10}                 \\ \hline
\multirow{2}{*}{\textbf{Models}}      & Teacher       & \multicolumn{1}{l}{LeNet5}      & \multicolumn{1}{l}{LeNet5}      & \multicolumn{1}{l}{AlexNet}         \\
                                      & Student       & \multicolumn{1}{l}{LeNet5 Half} & \multicolumn{1}{l}{LeNet5 Half} & \multicolumn{1}{l}{AlexNet Half}    \\ \hline
\multirow{2}{*}{\textbf{Baselines}}   & Normal  \cite{normal}      & 0.8091                           & 0.7591                           & 0.7332                            \\
                                      & Teacher Only  & 0.8051                           & 0.7604                           & 0.7076                           \\ \hline
\multirow{3}{*}{\textbf{Proposed Methods}} & CCKD-L        & 0.7940                           & 0.7815                           & 0.6808                           \\
                                      & CCKD-T        & 0.8326                           & 0.7941                           & 0.7168                           \\
                                      & CCKD-T + Reg. & 0.8709                           & 0.7825                           & 0.7984                           \\ \hline
\end{tabular}
\label{Table:9}
\caption{Effectiveness against adversarial samples.}
\end{table}

\section{Conclusions}
 The established approaches in distillation literature employ fixed loss functions and targets and are thus incapable of capturing the different levels of information present in different input data samples. CCKD-L and CCKD-T formulations are presented to address this issue. These methods employ sample specific loss functions and targets based on the teacher model's confidence. If the teacher's predictions are correct, then more weightage is assigned to supervision from the teacher and vice versa. These methods are response-based methods, and experiments on benchmark datasets (MNIST, Fashion MNIST, CIFAR10, and SVHN) establish their competitive performance against other state-of-the-art response-based KD methods. In most cases, the proposed methods outperform the other state-of-the-art methods. The data efficiency of CCKD methods is improved by incorporating self-regulation. Experiments on the MNIST dataset show that CCKD methods can achieve almost the same generalization performance as the other methods by using only $0.86\%$ of the total data samples. In addition to having better generalization capacity, the student models distilled through CCKD methods do not repeat the same mistakes as their teacher models on the training set. Compared to others, CCKD methods achieve an order of magnitude higher success rates and an order of magnitude lower failure rates, establishing their superiority for knowledge transfer. Following this finding, the performance of the CCKD methods against FGSM adversarial attacks was investigated. It was found that the student models distilled through these methods exhibited enhanced resistance to adversarial attacks. The proposed methods are at least $3\%$ better than the conventional distillation method \cite{normal} in resisting adversarial attacks which shows that they are better in transferring the robustness of the teacher models to the student models. This effect is demonstrated through experiments on the MNIST, Fashion MNIST, and CIFAR10 datasets. These beneficial features establish the superiority of the CCKD methods over other distillation methods.

As future prospects of this research, the extension of these formulations to an online, collaborative learning framework will be investigated. CCKD methods will also be extended to data-free distillations. Convergence characteristics and regularization properties of these methods will also be investigated. Other types of adversarial attacks and the performance of the CCKD methods against these will also be explored. Currently, only the robustness of the student models to adversarial samples is considered, and a comparison is made across the normal, teacher only, and the proposed distillation methods. Tranfer of adversarial samples from the teacher model to the student model will also be explored.

\bibliographystyle{elsarticle-num}
\bibliography{refs}

\end{document}